\documentclass{article}

\usepackage[main, final]{neurips_2025}
\usepackage{amsmath}
\usepackage{amssymb}
\usepackage{tabularx}
\usepackage{amsthm}
\usepackage{adjustbox}
\usepackage{algpseudocode}

\usepackage{mathtools}
\usepackage{array}
\usepackage{pdfpages}
\usepackage{multirow} 
\usepackage{wrapfig} 
\usepackage{colortbl}
\usepackage{subcaption}
\usepackage{graphicx}
\usepackage[ruled,linesnumbered]{algorithm2e}

\usepackage[utf8]{inputenc} % allow utf-8 input
\usepackage[T1]{fontenc}    % use 8-bit T1 fonts
\usepackage{hyperref}       % hyperlinks
\usepackage{url}            % simple URL typesetting
\usepackage{booktabs}       % professional-quality tables
\usepackage{amsfonts}       % blackboard math symbols
\usepackage{nicefrac}       % compact symbols for 1/2, etc.
\usepackage{microtype}      % microtypography
\usepackage{xcolor}         % colors

\title{HQA-VLAttack: Towards High Quality Adversarial Attack on Vision-Language Pre-Trained Models}

\author{
  Han Liu \\
  Dalian University of Technology \\ 
  Dalian, China \\
  \texttt{liu.han.dut@gmail.com} \\
  \And
  Jiaqi Li \\
  Dalian University of Technology \\
  Dalian, China \\
  \texttt{li.jiaqi.dut@gmail.com} \\
  \And
  Zhi Xu \\
  Dalian University of Technology \\
  Dalian, China \\
  \texttt{xu.zhi.dut@gmail.com} \\
  \And
  Xiaotong Zhang\thanks{Corresponding author.}\\
  Dalian University of Technology \\
  Dalian, China \\
  \texttt{zxt.dut@hotmail.com} \\
  \And
  Xiaoming Xu\\
  Dalian University of Technology \\
  Dalian, China \\
  \texttt{xu.xm.dut@gmail.com} \\
  \And
  Fenglong Ma \\
  The Pennsylvania State University \\
  Pennsylvania, USA \\
  \texttt{fenglong@psu.edu} \\
  \And
  Yuanman Li \\
  Shenzhen University \\
  Shenzhen, China \\
  \texttt{yuanmanli@szu.edu.cn} \\
  \And
  Hong Yu \\
  Dalian University of Technology \\
  Dalian, China \\
  \texttt{hongyu@dlut.edu.cn} \\
}

\begin{document}

\maketitle

\begin{abstract}
Black-box adversarial attack on vision-language pre-trained models is a practical and challenging task, as text and image perturbations need to be considered simultaneously, and only the predicted results are accessible. Research on this problem is in its infancy, and only a handful of methods are available. Nevertheless, existing methods either rely on a complex iterative cross-search strategy, which inevitably consumes numerous queries, or only consider reducing the similarity of positive image-text pairs but ignore that of negative ones, which will also be implicitly diminished, thus inevitably affecting the attack performance. To alleviate the above issues, we propose a simple yet effective framework to generate high-quality adversarial examples on vision-language pre-trained models, named HQA-VLAttack, which consists of text and image attack stages. For text perturbation generation, it leverages the counter-fitting word vector to generate the substitute word set, thus guaranteeing the semantic consistency between the substitute word and the original word. For image perturbation generation, it first initializes the image adversarial example via the layer-importance guided strategy, and then utilizes contrastive learning to optimize the image adversarial perturbation, which ensures that the similarity of positive image-text pairs is decreased while that of negative image-text pairs is increased. In this way, the optimized adversarial images and texts are more likely to retrieve negative examples, thereby enhancing the attack success rate. Experimental results on three benchmark datasets demonstrate that HQA-VLAttack significantly outperforms strong baselines in terms of attack success rate.
\end{abstract}

\section{Introduction}
Vision-Language Pre-training (VLP) models have become a cornerstone for cross-modal tasks, achieving remarkable success in applications such as image-text retrieval~\cite{itr2,itr1,itr3}, image captioning~\cite{IC}, and visual grounding \cite{DBLP:conf/eccv/MaJWYQ24}. However, research has shown that these models are vulnerable to adversarial attacks~\cite{SGA,DRA,Co-Attack,multimodalattack1,gao2023adversarial}, posing significant societal concerns. Adversarial attacks inject imperceptible perturbations to text and image inputs, aiming to manipulate predictions of victim VLP models maliciously. Specifically, existing attacks can be broadly categorized into white-box attacks~\cite{Co-Attack,PGD,white1_vl} and black-box attacks~\cite{SGA,DRA,VLAttack,BERT-ATTACK,black2_q}. In white-box attacks, attackers have full access to the victim model, allowing them to exploit gradients for highly effective attacks. However, the white-box setting can be too idealistic in real-world scenarios. In contrast, black-box attacks assume limited access to the victim model, such as confidence scores or prediction labels, making them more practical for real-world applications. 

Black-box attacks can be categorized into \textbf{query-based attacks}~\cite{VLAttack,BERT-ATTACK,black2_q, HQA-Attack, SSPAttack} and \textbf{transfer-based attacks}~\cite{SGA,DRA,VQAttack,UAP}. Query-based attacks employ an iterative cross-search strategy that requires repeatedly querying the victim model and utilizing its feedback to refine adversarial perturbations. While effective, these methods incur substantial query costs, limiting their practicality in real-world applications. In contrast, transfer-based attacks generate adversarial examples by optimizing them on a surrogate model, leveraging feature similarity and generalization to maintain their effectiveness against unseen victim models without requiring queries. Due to their independence from direct access to the victim model, transfer-based attacks are particularly well-suited for real-world adversarial scenarios, making their enhancement a critical research focus.

\begin{wrapfigure}{t}{0.45\textwidth}
    \centering
    \includegraphics[width=0.95\linewidth]{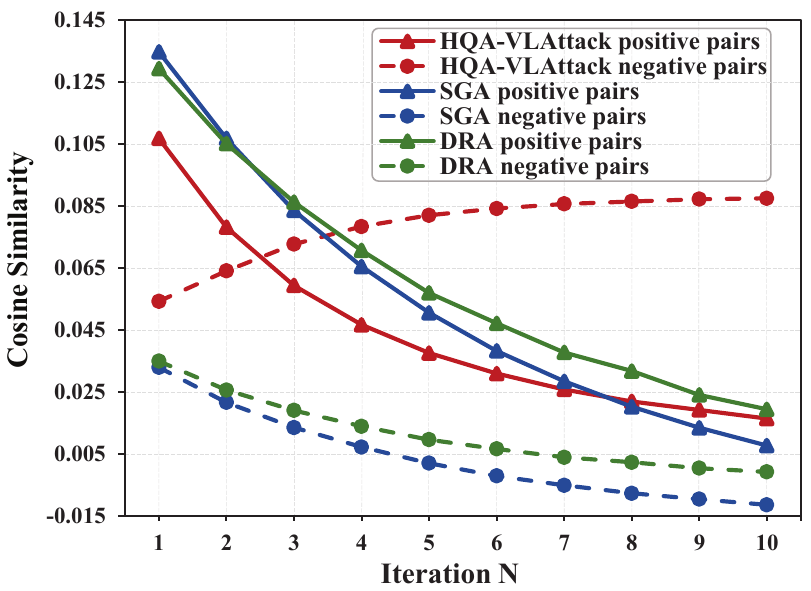}
       \caption{\small The average cosine similarity of image-text pairs optimized by SGA, DRA, and HQA-VLAttack on the Flickr30K dataset using ALBEF as the surrogate model.}
    \label{fig:iterator_problem}
     \vspace{-5pt}
\end{wrapfigure}

As shown in Figure~\ref{fig:iterator_problem}, existing transfer-based adversarial attacks on VLP models primarily aim to decrease the similarity of positive image-text pairs, thus causing the victim model to retrieve more negative examples and improving the attack success rates. Specifically, SGA~\cite{SGA} utilizes BERT-Attack for text perturbation along with set-level guidance to explicitly reduce the similarity between positive image-text pairs. In addition, DRA~\cite{DRA} enhances this approach by incorporating trajectory-aligned diversified sampling and text-guided selection, which maximizes the semantic distance between positive image-text pairs, further increasing the attack success rate. However, both SGA and DRA primarily focus on reducing the similarity of positive image-text pairs, inadvertently reducing the similarity of negative pairs as well. Consequently, VLP models may still be biased towards retrieving positive examples, which can inevitably diminish the overall attack success rates. For further experimental details regarding Figure~\ref{fig:iterator_problem}, please refer to Appendix A.

\begin{figure*}[t]
\begin{center}
   \includegraphics[width=0.95\linewidth]{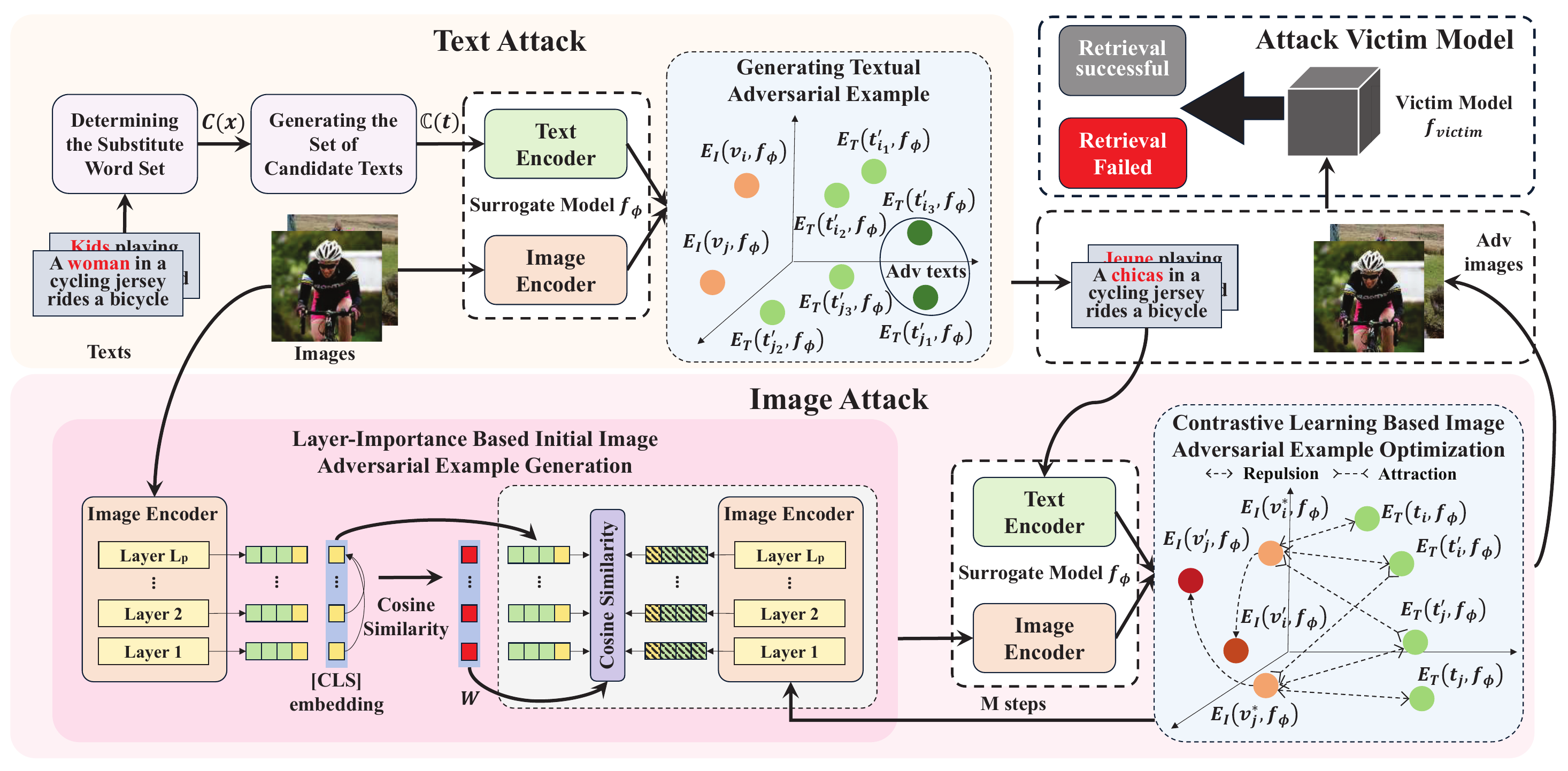}
\end{center}
  \vskip -0.2in
   \caption{The overall of HQA-VLAttack. First, the Text Attack module determines the substitute word set and then generates the textual adversarial example $t_i'$. Second, the Image Attack module applies layer-importance based initial image adversarial example generation to obtain the initial adversarial image example $v'_i$, followed by contrastive learning-based image adversarial example optimization for further refinement. Finally, the optimized adversarial examples are fed into the victim model.}
\label{fig:all}
\vspace{-15pt}
\end{figure*}

To address the limitation mentioned above, we propose a \textbf{H}igh \textbf{Q}uality transfer-based \textbf{A}dversarial \textbf{V}ision-\textbf{L}anguage \textbf{Attack}, namely \textbf{HQA-VLAttack}. The overview of HQA-VLAttack is depicted in Figure~\ref{fig:all}. By ``high quality'', we mean that HQA-VLAttack achieves a significantly higher attack success rate compared to existing methods. Specifically, HQA-VLAttack first generates semantically consistent adversarial texts, followed by adversarial images with low similarity to the original images. Finally, contrastive learning is used to increase the distance between matched adversarial image-text pairs while reducing the distance between unmatched pairs. Experimental results on three datasets and three attack tasks demonstrate that HQA-VLAttack outperforms other strong baselines, establishing it as an effective high-quality vision-language adversarial attack method\footnote{The source code is publicly available at \url{https://github.com/HQA-VLAttack/HQA-VLAttack}}.

\section{Related Work}
\subsection{White-Box Adversarial Attacks on VLP Models}
White-box adversarial attacks \cite{Co-Attack,white1_vl} assume that the attackers have full access to all information about the victim model, including its architecture, training data, and gradients. This enables attackers to directly exploit gradients and generate highly effective adversarial perturbations. Co-Attack \cite{Co-Attack} extends this paradigm to multimodal settings by simultaneously perturbing both image and text modalities, thereby leveraging intermodal dependencies to generate more effective adversarial examples. Its collaborative framework overcomes the limitations of single-modal attacks and significantly increases the attack success rate in various vision-language tasks. However, the white-box assumption can be too idealistic in real-world scenarios, as most developers will not release model details to the public. This significantly limits the practical deployment of such attack methods.

\subsection{Black-Box Adversarial Attacks on VLP Models}
Black-box adversarial attacks restrict access to limited model output, such as confidence scores or predicted labels, making them more practical for real-world applications. When targeting VLP models, these attacks can be broadly categorized into query-based and transfer-based attacks. 

\textbf{Query-based attacks} typically employ a complex iterative cross-search strategy on both image and text inputs, leading to high query consumption. VLAttack \cite{VLAttack} is one of the most advanced query-based attacks, achieving state-of-the-art performance in attack success rate. It generates adversarial examples by first independently perturbing images and texts, and then refining the adversarial pair through an iterative cross-search strategy that jointly optimizes the multimodal embedding. Although this method achieves effective attack performance, its reliance on a large number of queries significantly limits real-world applicability. 

\textbf{Transfer-based attacks} generate adversarial examples on a surrogate model and transfer them to deceive the victim model, thereby eliminating the need for queries. This makes transfer-based methods more practical for real-world applications. SGA \cite{SGA} leverages set-level attacks by generating adversarial examples from multi-scale images and multiple matching captions, thereby enhancing cross-modal interactions and transferability. DRA \cite{DRA} is a recent transfer-based method, which enhances transferability by diversifying adversarial examples along the intersection region of the adversarial trajectory and incorporates text-guided selection to mitigate overfitting. Most of these methods also use input transformation techniques~\cite{ID} to further improve transferability. However, current methods lack consideration for negative image-text pairs, causing biased retrieval in VLP models and limiting attack success rates. To address this issue, we propose a high-quality transfer-based attack that reduces the similarity of positive pairs while increasing the similarity of negative pairs. This approach makes VLP models more likely to retrieve negative examples, thereby improving attack performance.

\section{Problem Formulation}

In image-to-text retrieval (TR) task, the retrieval function $F_{TR}$ takes an input image $v_i$. The retrieval model $F_{TR}$ retrieves the top-$k$ candidate texts from a text set $D_t = \{t_1, \dots, t_n\}$, where $n$ denotes the set size. This retrieval process can be formulated as $F_{TR}(v_i, D_t)_k = \{t^{(1)}, \dots, t^{(k)}\}$.

In text-to-image retrieval (IR) task, the retrieval model $F_{IR}$ takes an input text $t_i = \{x_1, \dots, x_L\}$, where $L$ denotes the text length, and retrieves the top-$k$ candidate images from an image set $D_v = \{v_1, \dots, v_n\}$, where $n$ denotes the set size. Mathematically, this retrieval procedure can be expressed as $F_{IR}(t_i, D_v)_k = \{v_i^{(1)}, \dots, v_i^{(k)}\}$.

Transfer-based multimodal adversarial attacks aim to generate image and textual adversarial example sets, $D'_v = \{v'_1, \dots, v'_n\}$ and $D'_t = \{t'_1, \dots, t'_n\}$, by applying pixel-level perturbations to images and word-level perturbations to texts using a surrogate model $f_\phi$. Here, we use $(v'_i, t'_i)$ to represent a positive adversarial image-text pair. Then a successful attack on the TR task can be formulated as:
\begin{equation}
\label{eq:f_ir_define}
t'_i \notin F_{TR}(v'_i, D_t')_k \quad \text{s.t.} \quad \|v'_i - v_i\|_{\infty} \leq \epsilon_v, \, d(t'_i, t_i) \leq \epsilon_t,
\end{equation}
where $\|\cdot\|_\infty$ denotes the $L_{\infty}$ norm. The parameter $\epsilon_v$ represents the maximum allowable perturbation for images, while $d(\cdot, \cdot)$ measures the distance between adversarial and original texts, with $\epsilon_t$ denoting the maximum allowable perturbation for texts. Similarly, for the IR task:
\begin{equation}
\label{eq:attack_define}
v'_i \notin F_{IR}(t'_i, D_v')_k \quad \text{s.t.} \quad \|v'_i - v_i\|_{\infty} \leq \epsilon_v, \, d(t'_i, t_i) \leq \epsilon_t.
\end{equation}
These conditions ensure that adversarial examples effectively degrade retrieval performance.

\section{The Proposed Method}

\subsection{Text Attack} 
\subsubsection{Determining the Substitute Word Set}
To generate adversarial texts, we first create the semantically consistent substitute word sets, which are then used to replace words in the original text, ensuring effective perturbations that increase the likelihood of deceiving VLP models. Previous methods typically employed the masked language model (MLM) \cite{bert_mlm} to generate substitutes for specific positions in the text. However, MLM predictions rely solely on context, which can lead to semantically inconsistent words. For example, in the sentence “I [MASK] you.” [MASK] may be predicted as either “love” or “hate”. This inconsistency can severely degrade the effectiveness of generating adversarial text.

To address this issue, we use the counter-fitting word vector \cite{counter-fitting} to generate the substitute word set. Specifically, given a clean text $t_i=[x_1,...,x_j,...,x_L]$, where $x_j$ denotes the $j$-th word in the sentence, we generate a textual adversarial example $t'_i$ by replacing words in $t_i$. For each word $x_j$ in the sentence, if its corresponding word vector $\mathbf{v}_{x_j}$ exists in the counter-fitting word vector set $\mathbf{V}_{cf}$, we select all synonyms $x_j'$ whose word vectors $\mathbf{v}_{x_j'} \in \mathbf{V}_{cf}$ have a cosine similarity greater than $\tau$ with $\mathbf{v}_{x_j}$. If $\mathbf{v}_{x_j}$ does not exist in $\mathbf{V}_{cf}$, we generate $k$ synonyms using the method from BERT-Attack \cite{BERT-ATTACK}. This process can be formally defined as:
\begin{equation}
    \label{eq:subs_set}
    C(x_j)= \begin{cases} x_j' \mid \text{cos}(\mathbf{v}_{x_j'}, \mathbf{v}_{x_j}) > \tau, & \text{if} \; \mathbf{v}_{x_j} \in \mathbf{V}_{cf}, \\ 
    \text{argmax}_k f_{\text{mlm}}(x_j), & \text{otherwise},
\end{cases}
\end{equation}
where $C(x_j)$ denotes the set of substitute words for $x_j$, and $f_{mlm}(\cdot)$ represents the masked language model used in BERT-Attack to generate synonyms.

\subsubsection{Generating Textual Adversarial Example}
Intuitively, image-text pairs with lower similarity on the surrogate model tend to have lower similarity on the victim model as well, making these adversarial image-text pairs more likely to succeed in attacking the victim model. Therefore, to improve the attack success rate, we need to generate a textual adversarial example $t'_i$ by applying synonym replacement, such that its similarity with the corresponding image $v_i$ is lowest in the surrogate model. In several vision-language tasks, the text often consists of a limited number of tokens. Consequently, we restrict the replacement to a single word within the text to generate $t'_i$. Specifically, for each word $x_j$ in the text $t_i=[x_1,...,x_j,...,x_L]$, we iteratively substitute $x_j$ with each synonym from its substitute word set $C(x_j)$, thereby generating a set of candidate texts $\mathbb C({t_i})_j$. We then aggregate all candidate sets to form the comprehensive adversarial text collection $\mathbb C(t_i)$. The adversarial example $t'_i$ is subsequently selected based on the minimum cosine similarity to the original image, as formalized by the following equation:
\begin{equation}
    \label{eq:find_low_ti}
    t'_i = \underset{t^*_i \in \mathbb C(t_i)} {\operatorname{argmax}} -cos(E_{T}(t^*_i,f_\phi),E_{I}(v_i,f_\phi)),
\end{equation}
where $cos(\cdot,\cdot)$ is the cosine similarity function, $E_T(t^*_i,f_\phi)$ denotes the text feature of $t^*_i$ extracted by the text encoder of the surrogate model $f_\phi$, $E_I(v_i,f_\phi)$ denotes the image feature of $v_i$ extracted by the image encoder of the surrogate model $f_\phi$.

\subsection{Image Attack}

\subsubsection{Layer-Importance Based Initial Image Adversarial Example Generation}
This step aims to generate an initial image adversarial example by minimizing its similarity to the original image. Existing methods \cite{VLAttack,VQAttack} extract layer-wise representations and minimize their similarity iteratively. However, they wrongly assume equal layer contributions, while actual influence varies. To address this, we propose a layer-importance based method to generate an initial image adversarial example. This method involves two steps: determining layer importance and generating an initial image adversarial example.

\begin{wrapfigure}{t}{0.45\textwidth}
    
    \centering
    \includegraphics[width=0.95\linewidth]{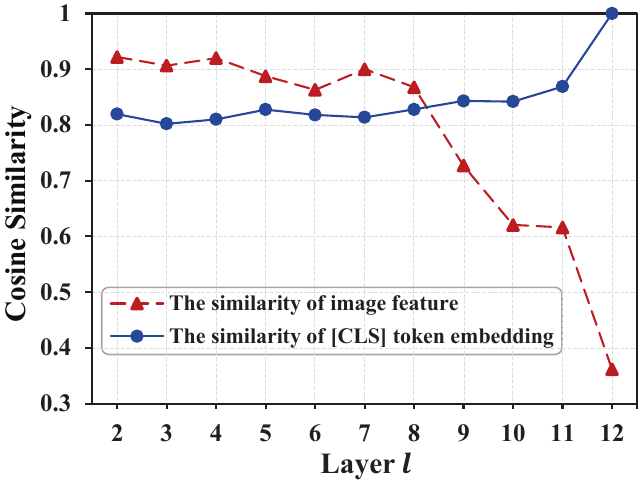}
       \caption{The cosine similarity of image feature and [CLS] token embedding across Layers.}
    \label{fig:layer-importance}
   \vspace{-10pt}
\end{wrapfigure}

\textbf{Determining layer importance.} We conduct an experiment to quantify the contribution of each layer in the model. As illustrated in Figure~\ref{fig:layer-importance}, we present two similarity variation curves: (1) the cosine similarity between the output feature of an image $v_i$ under normal propagation, denoted as $E_I(v_i, f_\phi)$, and the output feature obtained after skipping the $l$-th layer, denoted as $E_{I \backslash l}(v_i, f_\phi)$; (2) the cosine similarity between the [CLS] token embedding at the $l$-th layer, $E_I(v_i, f_\phi)_{l,1}$, and the [CLS] token embedding at the top layer $L_p$, $E_I(v_i, f_\phi)_{L_p,1}$. Further experimental details are provided in Appendix B. The results indicate that as the layer index increases, skipping a layer exerts a more pronounced influence on the final output, highlighting the greater significance of higher layers. Similarly, the similarity trend of the [CLS] token follows a comparable pattern, exhibiting a smoother variation. Since different training strategies result in varying parameter sensitivities, more substantial changes are observed in the top layers. Assigning higher importance weights to higher layers can lead to adversarial examples that overfit the surrogate model, thereby reducing their effectiveness across different models. The smooth variation observed in the [CLS] token suggests that this issue can be mitigated by narrowing the importance weight gap between lower and higher layers. Consequently, employing the cosine similarity between $E_I(v_i, f_\phi)_{l,1}$ and $E_I(v_i, f_\phi)_{L_p,1}$ as the importance weight is a reasonable choice. Thus, the $l$-th layer importance weight is defined as:  
\begin{equation}
\label{eq:layer-importance}
    w_{i,l} = cos(E_I(v_i,f_\phi)_{l,1},E_I(v_i,f_\phi)_{L_p,1}).
\end{equation}
In Eq.~\eqref{eq:layer-importance}, a higher value of $w_{i,l}$ indicates greater layer importance in generating adversarial examples.  

\textbf{Generate initial image adversarial example.} Given the computed layer importance, we optimize an adversarial image that differs significantly from the original in important layers by minimizing the weighted sum of feature similarities. Specifically, for an original image $v_i$, we introduce a random perturbation $\delta \sim U(-\epsilon,\epsilon)$ to obtain $v_i' = v_i + \delta$. We then compute the cosine similarity between the features $E_I(v_i,f_\phi)_{l,j}$ and $E_I(v'_i,f_\phi)_{l,j}$ extracted from the $j$-th token at the $l$-th layer of the surrogate model for both $v_i$ and $v'_i$. This optimization is formulated as:  
\begin{equation}
    \mathcal{L}_l = \sum_{l=1}^{L_p} w_{i,l} \times \frac{1}{D_p} \sum_{j=1}^{D_p}  \cos(E_I(v_i,f_\phi)_{l,j}, E_I(v'_i,f_\phi)_{l,j}),
\end{equation}  
where $D_p$ denotes the number of tokens per layer. Minimizing $\mathcal{L}_l$ reduces the similarity between the adversarial image $v'_i$ and the original image $v_i$. Finally, we apply PGD \cite{PGD} to optimize this objective and generate the initial image adversarial example $v'_i$.

\subsubsection{Contrastive Learning Based Image Adversarial Example Optimization}
To further reduce the similarity between positive image-text pairs and increase the similarity between negative image-text pairs—thereby encouraging both the surrogate and victim models to retrieve unmatched texts for $v'_i$ and ultimately enhancing the attack success rate—we employ a contrastive learning approach to optimize the adversarial image $v'_i$.

Specifically, let $T_p$ denote the set of textual adversarial examples generated from the texts matched with $v'_i$ in a batch of image-text pairs, along with their corresponding original texts. Let $T_n$ denote the set of adversarial texts that are unmatched with $v'_i$ within the same batch. During the optimization process, we design the following loss function:
\begin{equation}
\label{eq:loss_c}
\begin{split}
    \mathcal{L}_c = \sum_{v^*_i \in \text{Trans}(v'_i)} ( \lambda \sum_{t'_i \in T_p} \text{cos}(E_T(t'_i,f_\phi), E_I(v^*_i,f_\phi)) 
+ \sum_{t'_j \in T_{n}} \text{cos}(E_T(t'_j,f_\phi), E_I(v^*_i,f_\phi) ),
\end{split}
\end{equation}
where $\text{Trans}(v'_i)$ is the scale transformation function, and $\lambda$ is the penalty factor for positive image-text pairs in contrastive learning.

By means of this loss function, on the surrogate model $f_\phi$, for positive image-text pairs, we minimize their similarity, that is, increase their distance in the feature space; for negative image-text pairs, we maximize the similarity of their feature vectors, that is, reduce the distance between the two in the feature space. As a result, the optimized adversarial image example is more likely to retrieve negative texts, thereby improving the attack success rate. Finally, we utilize the PGD to optimize $\mathcal{L}_c$ and obtain the refined image adversarial example $v'_i$.

\subsection{The Overall Procedure}
HQA-VLAttack begins by extracting a batch of image-text pairs from the datasets $D_t$ and \( D_v \) in each round and optimizing them through the following procedure to generate \( D_t' \) and \( D_v' \). The process starts with a text attack: given an original image-text pair \( (v_i, t_i) \), HQA-VLAttack first determines the substitute word set for each candidate word in $t_i$. Using these substitute sets, it constructs the textual adversarial example \( t_i' \). Following this, an iterative image attack is conducted. In each iteration, HQA-VLAttack first initializes the image adversarial example $v_i'$, then uses a contrastive learning-based method to further optimize the image adversarial example. The detailed algorithm procedure of HQA-VLAttack is given in Appendix C.

\section{Experiment}
\subsection{Experimental Settings}

\textbf{Dataset.} We conduct experiments on three widely-used public multimodal datasets Flickr30K~\cite{Flickr30k}, MSCOCO~\cite{MSCOCO}, and RefCOCO+~\cite{Refcoco+}. The detailed dataset description is shown in Appendix D.
For the image-text retrieval task, we conduct experiments on the Flickr30K test set, which contains 1,000 images and 5,000 captions, as well as on the MSCOCO test set, which includes 5,000 images and approximately 25,000 captions. For visual grounding and image captioning tasks, we use 3,000 images and 15,000 captions from RefCOCO+ as well as 10,000 images and 50,000 captions from MSCOCO, respectively. We adopt the Karpathy split for experimental evaluation.

\begin{table*}[!t]
\centering
% \setlength{\tabcolsep}{3.8mm}
% \scalebox{1}{
\caption{\textbf{Attack success rate (\%) in image-text retrieval on the Flickr30K dataset.} We present the attack success rate metric R@1 for both IR and TR tasks, indicating the success of attacks at Rank 1. $^*$ indicates white-box attacks.}
\resizebox{\textwidth}{!}{
\begin{tabular}{@{}llcccccccc@{}}
\toprule
\multicolumn{10}{c}{\textbf{Flickr30K Dataset}} \\ \midrule
\multicolumn{1}{c|}{\multirow{2}{*}{\textbf{~Surrogate Model~}}}   & \multicolumn{1}{c|}{\textbf{Victim Model}}  & \multicolumn{2}{c|}{\textbf{ALBEF}}                  & \multicolumn{2}{c|}{\textbf{TCL}}                    & \multicolumn{2}{c|}{\textbf{CLIP\textsubscript{\textbf{ViT}}}}               & \multicolumn{2}{c}{\textbf{CLIP\textsubscript{\textbf{CNN}}}} \\ \cmidrule(l){2-10} 
\multicolumn{1}{c|}{}                                   & \multicolumn{1}{c|}{\textbf{Attack Method}}  & ~TR R@1~         & \multicolumn{1}{c|}{~IR R@1~}         & ~TR R@1~         & \multicolumn{1}{c|}{~IR R@1~}         & ~TR R@1~         & \multicolumn{1}{c|}{~IR R@1~}         & ~TR R@1~            & ~IR R@1~            \\ \midrule
\multicolumn{1}{l|}{\multirow{8}{*}{\textbf{ALBEF}}}    & \multicolumn{1}{l|}{PGD}              & 52.45$^*$          & \multicolumn{1}{c|}{58.65$^*$}          & 3.06           & \multicolumn{1}{c|}{6.79}           & 8.96           & \multicolumn{1}{c|}{13.21}          & 10.34             & 14.65             \\
\multicolumn{1}{l|}{}                                   & \multicolumn{1}{l|}{BERT-Attack}      & 11.57$^*$          & \multicolumn{1}{c|}{27.46$^*$}          & 12.64          & \multicolumn{1}{c|}{28.07}          & 29.33          & \multicolumn{1}{c|}{43.17}          & 32.69             & 46.11             \\
\multicolumn{1}{l|}{}                                   & \multicolumn{1}{l|}{Sep-Attack}       & 65.69$^*$          & \multicolumn{1}{c|}{73.95$^*$}          & 17.60          & \multicolumn{1}{c|}{32.95}          & 31.17          & \multicolumn{1}{c|}{45.23}         & 32.83             & 45.49             \\
\multicolumn{1}{l|}{}                                   & \multicolumn{1}{l|}{Co-Attack}        & 77.16$^*$          & \multicolumn{1}{c|}{83.86$^*$}          & 15.21          & \multicolumn{1}{c|}{29.49}          & 23.60          & \multicolumn{1}{c|}{36.48}          & 25.12             & 38.89             \\
\multicolumn{1}{l|}{}                                   & \multicolumn{1}{l|}{SGA}              & 97.24$^*$ & \multicolumn{1}{c|}{97.28$^*$}          & 45.42          & \multicolumn{1}{c|}{55.25}          & 33.38          & \multicolumn{1}{c|}{44.16}          & 34.93             & 46.57             \\
\multicolumn{1}{l|}{}                                   
& \multicolumn{1}{l|}{DRA}              & 96.14$^*$ & \multicolumn{1}{c|}{96.63$^*$}          & 49.74          & \multicolumn{1}{c|}{58.83}          & 39.14          & \multicolumn{1}{c|}{48.39}          &41.38             & 51.66             \\
\multicolumn{1}{l|}{}       &
\multicolumn{1}{l|}{\textbf{HQA-VLAttack}}  & {\textbf{99.79$^\ast$}} & \multicolumn{1}{c|}{\textbf{99.98$^\ast$}} &{\textbf{73.02}}& \multicolumn{1}{c|}{\textbf{77.60}} & {\textbf{52.15}} & \multicolumn{1}{c|}{\textbf{62.05}} & {\textbf{59.64}} & {\textbf{65.59}} \\
\midrule
\multicolumn{1}{l|}{\multirow{6}{*}{\textbf{TCL}}}      & \multicolumn{1}{l|}{PGD}              & 6.15           & \multicolumn{1}{c|}{10.78}          & 77.87$^*$          & \multicolumn{1}{c|}{79.48$^*$}          & 7.48           & \multicolumn{1}{c|}{13.72}          & 10.34             & 15.33             \\
\multicolumn{1}{l|}{}                                   & \multicolumn{1}{l|}{BERT-Attack}      & 11.89          & \multicolumn{1}{c|}{26.82}          & 14.54$^*$          & \multicolumn{1}{c|}{29.17$^*$}          & 29.69          & \multicolumn{1}{c|}{44.49}          & 33.46             & 46.06             \\
\multicolumn{1}{l|}{}                                   & \multicolumn{1}{l|}{Sep-Attack}       & 20.13          & \multicolumn{1}{c|}{36.48}          & 84.72$^*$          & \multicolumn{1}{c|}{86.07$^*$}          & 31.29          & \multicolumn{1}{c|}{44.65}          & 33.33             & 45.80             \\
\multicolumn{1}{l|}{}                                   & \multicolumn{1}{l|}{Co-Attack}        & 23.15          & \multicolumn{1}{c|}{40.04}          & 77.94$^*$          & \multicolumn{1}{c|}{85.59$^*$}          & 27.85          & \multicolumn{1}{c|}{41.19}          & 30.74             & 44.11             \\
\multicolumn{1}{l|}{}                                   & \multicolumn{1}{l|}{SGA}              & 48.91          & \multicolumn{1}{c|}{60.34}          & 98.37$^*$ & \multicolumn{1}{c|}{98.81$^*$} & 33.87          & \multicolumn{1}{c|}{44.88}          & 37.74             & 48.30             \\
\multicolumn{1}{l|}{}                                   
& \multicolumn{1}{l|}{DRA}              & 51.09 & \multicolumn{1}{c|}{61.79}          & 98.21$^*$        & \multicolumn{1}{c|}{98.33$^*$}          &40.25          & \multicolumn{1}{c|}{48.94}          &42.91            & 52.49            \\
\multicolumn{1}{l|}{}                                   
& \multicolumn{1}{l|}{ \textbf{HQA-VLAttack}} & {\textbf{62.88}} & \multicolumn{1}{c|}{ {\textbf{71.70}}} & {\textbf{99.79$^*$}}          & \multicolumn{1}{c|}{ {\textbf{99.93$^*$}} }         & {\textbf{52.39}} & \multicolumn{1}{c|}{ {\textbf{59.41}}} & {\textbf{55.43}}    & {\textbf{62.44}}    \\ \midrule
\multicolumn{1}{l|}{\multirow{6}{*}{\textbf{CLIP\textsubscript{\textbf{ViT}}}}} & \multicolumn{1}{l|}{PGD}              & 2.50           & \multicolumn{1}{c|}{4.93}           & 4.85           & \multicolumn{1}{c|}{8.17}           & 70.92$^*$          & \multicolumn{1}{c|}{78.61$^*$}          & 5.36              & 8.44              \\
\multicolumn{1}{l|}{}                                   & \multicolumn{1}{l|}{BERT-Attack}      & 9.59           & \multicolumn{1}{c|}{22.64}          & 11.80          & \multicolumn{1}{c|}{25.07}          & 28.34$^*$          & \multicolumn{1}{c|}{39.08$^*$}          & 30.40             & 37.43             \\
\multicolumn{1}{l|}{}                                   & \multicolumn{1}{l|}{Sep-Attack}       & 9.59           & \multicolumn{1}{c|}{23.25}          & 11.38          & \multicolumn{1}{c|}{25.60}          & 79.75$^*$          & \multicolumn{1}{c|}{86.79$^*$}          & 30.78             & 39.76             \\
\multicolumn{1}{l|}{}                                   & \multicolumn{1}{l|}{Co-Attack}        & 10.57          & \multicolumn{1}{c|}{24.33}          & 11.94          & \multicolumn{1}{c|}{26.69}          & 93.25$^*$          & \multicolumn{1}{c|}{95.86$^*$}          & 32.52             & 41.82             \\
\multicolumn{1}{l|}{}                                   & \multicolumn{1}{l|}{SGA}              & 13.40          & \multicolumn{1}{c|}{27.22}          & 16.23          & \multicolumn{1}{c|}{30.76}          & 99.08$^*$ & \multicolumn{1}{c|}{98.94$^*$} & 38.76             & 47.79             \\
\multicolumn{1}{l|}{}                                   
& \multicolumn{1}{l|}{DRA}              & 12.51 & \multicolumn{1}{c|}{30.00}          & 14.65        & \multicolumn{1}{c|}{30.62}          &98.77$^*$       & \multicolumn{1}{c|}{99.00$^*$}          &45.47            & 50.74            \\
\multicolumn{1}{l|}{}                                   
& \multicolumn{1}{l|}{  \textbf{HQA-VLAttack}} & {\textbf{25.13}} & \multicolumn{1}{c|}{ {\textbf{41.98}}} & {\textbf{24.66}} & \multicolumn{1}{c|}{ {\textbf{44.00}}} & {\textbf{100.00$^*$}}          & \multicolumn{1}{c|}{ {\textbf{100.00$^*$}}}          & {\textbf{74.07}}    & {\textbf{77.19}}    \\ \midrule
\multicolumn{1}{l|}{\multirow{6}{*}{\textbf{CLIP\textsubscript{\textbf{CNN}}}}} & \multicolumn{1}{l|}{PGD}              & 2.09           & \multicolumn{1}{c|}{4.82}           & 4.00           & \multicolumn{1}{c|}{7.81}           & 1.10           & \multicolumn{1}{c|}{6.60}           & 86.46$^*$             & 92.25$^*$             \\
\multicolumn{1}{l|}{}                                   & \multicolumn{1}{l|}{BERT-Attack}      & 8.86           & \multicolumn{1}{c|}{23.27}          & 12.33          & \multicolumn{1}{c|}{25.48}          & 27.12          & \multicolumn{1}{c|}{37.44}          & 30.40$^*$             & 40.10$^*$             \\
\multicolumn{1}{l|}{}                                   & \multicolumn{1}{l|}{Sep-Attack}       & 8.55           & \multicolumn{1}{c|}{23.41}          & 12.64          & \multicolumn{1}{c|}{26.12}          & 28.34          & \multicolumn{1}{c|}{39.43}          & 91.44$^*$             & 95.44$^*$             \\
\multicolumn{1}{l|}{}                                   & \multicolumn{1}{l|}{Co-Attack}        & 8.79           & \multicolumn{1}{c|}{23.74}          & 13.10          & \multicolumn{1}{c|}{26.07}          & 28.79          & \multicolumn{1}{c|}{40.03}          & 94.76$^*$             & 96.89$^*$             \\
\multicolumn{1}{l|}{}                                   & \multicolumn{1}{l|}{SGA}              & 11.42          & \multicolumn{1}{c|}{24.80}          & 14.91          & \multicolumn{1}{c|}{28.82}          & 31.24          & \multicolumn{1}{c|}{42.12}          & 99.24$^*$             & 99.49$^*$             \\
\multicolumn{1}{l|}{}                                   
& \multicolumn{1}{l|}{DRA}              & 12.20 & \multicolumn{1}{c|}{26.59}          & 14.33        & \multicolumn{1}{c|}{29.29}          &35.21       & \multicolumn{1}{c|}{45.94}          &99.11$^*$           & 99.49$^*$            \\
\multicolumn{1}{l|}{}                                   
& \multicolumn{1}{l|}{ \textbf{HQA-VLAttack}} & {\textbf{20.75}} & \multicolumn{1}{c|}{ {\textbf{38.66}}} & {\textbf{22.13}} & \multicolumn{1}{c|}{ {\textbf{42.45}}} & {\textbf{62.82}} & \multicolumn{1}{c|}{ {\textbf{69.46}}} & {\textbf{99.87$^*$}}   & {\textbf{100.00$^*$}}    \\ \bottomrule
\end{tabular}
}
\vspace{-20pt}
\label{tab:flickr-ITR}
\end{table*}

\textbf{Models.} We follow~\cite{SGA,DRA} to evaluate two popular VLP architectures, the fused VLP and aligned VLP models. Specifically, we select ALBEF~\cite{ALBEF} and TCL~\cite{TCL} as representatives of the fused VLP category. ALBEF integrates a 12-layer visual transformer (ViT-B/16)~\cite{ViT} as the image encoder, and employs two 6-layer transformers as the text encoder and multimodal encoder, respectively. TCL shares the same architectural framework as ALBEF but is distinguished by its unique pre-training objectives. For the aligned VLP model, we focus on CLIP~\cite{CLIP}, which offers two distinct image encoder variants: $\mathrm{CLIP_{ViT}}$ and $\mathrm{CLIP_{CNN}}$. These variants leverage ViT-B/16 and ResNet-101~\cite{resnet} as their respective base architectures for the image encoder.

\textbf{Baselines.} We compare HQA-VLAttack with the following baselines: (1) \textbf{PGD}~\cite{PGD} is a white-box image adversarial attack method that iteratively maximizes model loss under perturbation constraints via projected gradient descent. (2) \textbf{BERT-Attack}~\cite{BERT-ATTACK} is a black-box query-based textual adversarial attack method that crafts context-aware substitutions via BERT to fool NLP models with minimal edits. (3) \textbf{Sep-Attack}~\cite{SGA} is a black-box transfer-based multimodal adversarial attack method that separately perturbs unimodal data without any cross-modal interactions. (4) \textbf{Co-Attack}~\cite{Co-Attack} is a white-box multimodal adversarial attack method that collaboratively perturbs both image and text modalities to enhance adversarial effects. (5) \textbf{SGA}~\cite{SGA} is a black-box transfer-based multimodal adversarial attack method that uses set-level attacks to boost adversarial transferability in vision-language models. (6) \textbf{DRA}~\cite{DRA} is a recent black-box transfer-based multimodal adversarial attack method that enhances adversarial transferability by diversifying adversarial examples along the intersection region of the adversarial trajectory.

\begin{table*}[h]
\centering
\caption{\textbf{Attack success rate (\%) in image-text retrieval on the MSCOCO Dataset.} We present the attack success rate metric R@1 for both IR and TR tasks, indicating the success of attacks at Rank 1. $^*$ indicates white-box attacks.}
\resizebox{\textwidth}{!}{
\begin{tabular}{@{}llcccccccc@{}}
\toprule
\multicolumn{10}{c}{\textbf{MSCOCO Dataset}}                                                                                                                                                                                                                                                \\ \midrule
\multicolumn{1}{c|}{\multirow{2}{*}{\textbf{~Surrogate Model~}}}   & \multicolumn{1}{c|}{\textbf{Victim Model}}      & \multicolumn{2}{c|}{\textbf{ALBEF}}                           & \multicolumn{2}{c|}{\textbf{TCL}}                             & \multicolumn{2}{c|}{\textbf{CLIP\textsubscript{\textbf{ViT}}}}                        & \multicolumn{2}{c}{\textbf{CLIP\textsubscript{\textbf{CNN}}}}    \\ \cmidrule(l){2-10} 
\multicolumn{1}{c|}{}                          & \multicolumn{1}{c|}{\textbf{Attack Method}}      & ~TR R@1~         & \multicolumn{1}{c|}{~IR R@1~}         & ~TR R@1~         & \multicolumn{1}{c|}{~IR R@1~}         & ~TR R@1~         & \multicolumn{1}{c|}{~IR R@1~}         & ~TR R@1~         & ~IR R@1~         \\ \midrule
\multicolumn{1}{l|}{\multirow{6}{*}{\textbf{ALBEF}}}    & \multicolumn{1}{l|}{PGD}         & 76.70$^*$          & \multicolumn{1}{c|}{86.30$^*$}          & 12.46          & \multicolumn{1}{c|}{17.77}          & 13.96          & \multicolumn{1}{c|}{23.10}          & 17.45          & 23.54          \\
\multicolumn{1}{l|}{}                          & \multicolumn{1}{l|}{BERT-Attack} & 24.39$^*$          & \multicolumn{1}{c|}{36.13$^*$}          & 24.34          & \multicolumn{1}{c|}{33.39}          & 44.94          & \multicolumn{1}{c|}{52.28}          & 47.73          & 54.75          \\
\multicolumn{1}{l|}{}                          & \multicolumn{1}{l|}{Sep-Attack}  & 82.60$^*$          & \multicolumn{1}{c|}{89.88$^*$}          & 32.83          & \multicolumn{1}{c|}{42.92}          & 44.03          & \multicolumn{1}{c|}{54.46}          & 46.96          & 55.88          \\
\multicolumn{1}{l|}{}                          & \multicolumn{1}{l|}{Co-Attack}   & 79.87$^*$          & \multicolumn{1}{c|}{87.83$^*$}          & 32.62          & \multicolumn{1}{c|}{43.09}          & 44.89          & \multicolumn{1}{c|}{54.75}          & 47.30          & 55.64          \\
\multicolumn{1}{l|}{}                          & \multicolumn{1}{l|}{SGA}         & 96.75$^*$          & \multicolumn{1}{c|}{96.95$^*$}          & 58.56          & \multicolumn{1}{c|}{65.38}          & 57.06          & \multicolumn{1}{c|}{65.25}          & 58.95          & 66.52          \\
\multicolumn{1}{l|}{}                          & \multicolumn{1}{l|}{DRA}         & 96.57$^*$          & \multicolumn{1}{c|}{96.47$^*$}          & 60.69         & \multicolumn{1}{c|}{67.46}          & 61.69          & \multicolumn{1}{c|}{67.43}          &62.32          & 69.22          \\
\multicolumn{1}{l|}{}                          & \multicolumn{1}{l|}{\textbf{HQA-VLAttack}}        & \textbf{99.97$^*$} & \multicolumn{1}{c|}{\textbf{99.97$^*$}} & \textbf{85.69} & \multicolumn{1}{c|}{\textbf{87.81}} & \textbf{75.35} & \multicolumn{1}{c|}{\textbf{80.20}} & \textbf{77.52} & \textbf{81.64} \\ \midrule
\multicolumn{1}{l|}{\multirow{6}{*}{\textbf{TCL}}}      & \multicolumn{1}{l|}{PGD}         & 10.83          & \multicolumn{1}{c|}{16.52}          & 59.58$^*$          & \multicolumn{1}{c|}{69.53$^*$}          & 14.23          & \multicolumn{1}{c|}{22.28}          & 17.25          & 23.12          \\
\multicolumn{1}{l|}{}                          & \multicolumn{1}{l|}{BERT-Attack} & 35.32          & \multicolumn{1}{c|}{45.92}          & 38.54$^*$          & \multicolumn{1}{c|}{48.48$^*$}          & 51.09          & \multicolumn{1}{c|}{58.80}          & 52.23          & 61.26          \\
\multicolumn{1}{l|}{}                          & \multicolumn{1}{l|}{Sep-Attack}  & 41.71          & \multicolumn{1}{c|}{52.97}          & 70.32$^*$          & \multicolumn{1}{c|}{78.97$^*$}          & 50.74          & \multicolumn{1}{c|}{60.13}          & 51.90          & 61.26          \\
\multicolumn{1}{l|}{}                          & \multicolumn{1}{l|}{Co-Attack}   & 46.08          & \multicolumn{1}{c|}{57.09}          & 85.38$^*$          & \multicolumn{1}{c|}{91.39$^*$}          & 51.62          & \multicolumn{1}{c|}{60.46}          & 52.13          & 62.49          \\
\multicolumn{1}{l|}{}                          & \multicolumn{1}{l|}{SGA}         & 65.93          & \multicolumn{1}{c|}{73.30}           & 98.97$^*$ & \multicolumn{1}{c|}{99.15$^*$}          & 56.34          & \multicolumn{1}{c|}{63.99}          & 59.44          & 65.70           \\
\multicolumn{1}{l|}{}                          & \multicolumn{1}{l|}{DRA}         & 68.06          & \multicolumn{1}{c|}{75.86}           & 98.99$^*$ & \multicolumn{1}{c|}{99.15$^*$}          & 63.30          & \multicolumn{1}{c|}{63.99}          & 64.24          & 65.70           \\
\multicolumn{1}{l|}{}                          & \multicolumn{1}{l|}{\textbf{HQA-VLAttack}}        & \textbf{81.90} & \multicolumn{1}{c|}{\textbf{86.32}} & \textbf{99.97$^*$}          & \multicolumn{1}{c|}{\textbf{99.97$^*$}} & \textbf{70.77} & \multicolumn{1}{c|}{\textbf{76.22}} & \textbf{73.52} & \textbf{78.22} \\ \midrule
\multicolumn{1}{l|}{\multirow{6}{*}{\textbf{CLIP\textsubscript{\textbf{ViT}}}}} & \multicolumn{1}{l|}{PGD}         & 7.24           & \multicolumn{1}{c|}{10.75}          & 10.19          & \multicolumn{1}{c|}{13.74}          & 54.79$^*$          & \multicolumn{1}{c|}{66.85$^*$}          & 7.32           & 11.34          \\
\multicolumn{1}{l|}{}                          & \multicolumn{1}{l|}{BERT-Attack} & 20.34          & \multicolumn{1}{c|}{29.74}          & 21.08          & \multicolumn{1}{c|}{29.61}          & 45.06$^*$          & \multicolumn{1}{c|}{51.68$^*$}          & 44.54          & 55.32          \\
\multicolumn{1}{l|}{}                          & \multicolumn{1}{l|}{Sep-Attack}  & 23.41          & \multicolumn{1}{c|}{34.61}          & 25.77          & \multicolumn{1}{c|}{36.84}          & 68.52$^*$          & \multicolumn{1}{c|}{77.94$^*$}          & 43.11          & 49.76          \\
\multicolumn{1}{l|}{}                          & \multicolumn{1}{l|}{Co-Attack}   & 30.28          & \multicolumn{1}{c|}{42.67}          & 32.84          & \multicolumn{1}{c|}{44.69}          & 97.98$^*$          & \multicolumn{1}{c|}{98.80$^*$}          & 55.08          & 62.51          \\
\multicolumn{1}{l|}{}                          & \multicolumn{1}{l|}{SGA}         & 33.41          & \multicolumn{1}{c|}{44.64}          & 37.54          & \multicolumn{1}{c|}{47.76}          & 99.79$^*$ & \multicolumn{1}{c|}{99.79$^*$} & 58.93          & 65.83          \\
\multicolumn{1}{l|}{}                          & \multicolumn{1}{l|}{DRA}         & 35.96          & \multicolumn{1}{c|}{48.00}          & 36.32          & \multicolumn{1}{c|}{48.56}          & 99.66$^*$ & \multicolumn{1}{c|}{99.70$^*$} & 64.41          & 69.99          \\
\multicolumn{1}{l|}{}                          & \multicolumn{1}{l|}{\textbf{HQA-VLAttack}}        & \textbf{56.05} & \multicolumn{1}{c|}{\textbf{66.03}} & \textbf{54.10} & \multicolumn{1}{c|}{\textbf{65.25}} & \textbf{100.00$^*$}          & \multicolumn{1}{c|}{\textbf{100.00$^*$}}          & \textbf{88.93} & \textbf{89.26} \\ \midrule
\multicolumn{1}{l|}{\multirow{6}{*}{\textbf{CLIP\textsubscript{\textbf{CNN}}}}} & \multicolumn{1}{l|}{PGD}         & 7.01           & \multicolumn{1}{c|}{10.62}          & 10.08          & \multicolumn{1}{c|}{13.65}          & 4.88           & \multicolumn{1}{c|}{10.70}          & 76.99$^*$          & 84.20$^*$          \\
\multicolumn{1}{l|}{}                          & \multicolumn{1}{l|}{BERT-Attack} & 23.38          & \multicolumn{1}{c|}{34.64}          & 24.58          & \multicolumn{1}{c|}{29.61}          & 51.28          & \multicolumn{1}{c|}{57.49}          & 54.43$^*$          & 62.17$^*$          \\
\multicolumn{1}{l|}{}                          & \multicolumn{1}{l|}{Sep-Attack}  & 26.53          & \multicolumn{1}{c|}{39.29}          & 30.26          & \multicolumn{1}{c|}{41.51}          & 50.44          & \multicolumn{1}{c|}{57.11}          & 88.72$^*$          & 92.49$^*$          \\
\multicolumn{1}{l|}{}                          & \multicolumn{1}{l|}{Co-Attack}   & 29.83          & \multicolumn{1}{c|}{41.97}          & 32.97          & \multicolumn{1}{c|}{43.72}          & 53.10          & \multicolumn{1}{c|}{58.90}          & 96.72$^*$          & 98.56$^*$          \\
\multicolumn{1}{l|}{}                          & \multicolumn{1}{l|}{SGA}         & 31.61          & \multicolumn{1}{c|}{43.00}          & 34.81          & \multicolumn{1}{c|}{45.95}          & 56.62          & \multicolumn{1}{c|}{60.77}          & 99.61$^*$          & 99.80$^*$           \\
\multicolumn{1}{l|}{}                          & \multicolumn{1}{l|}{DRA}         & 33.26          & \multicolumn{1}{c|}{45.15}          & 33.89          & \multicolumn{1}{c|}{46.49}          & 59.60          & \multicolumn{1}{c|}{64.87}          & 99.51$^*$          & 99.70$^*$           \\
\multicolumn{1}{l|}{}                          & \multicolumn{1}{l|}{\textbf{HQA-VLAttack}}        & \textbf{52.20} & \multicolumn{1}{c|}{\textbf{62.13}} & \textbf{51.11} & \multicolumn{1}{c|}{\textbf{62.49}} & \textbf{82.72} & \multicolumn{1}{c|}{\textbf{84.56}} & \textbf{100.00$^*$} & \textbf{100.00$^*$} \\ \bottomrule
\end{tabular}
}
% \vspace{-3pt}
\vspace{-10pt}
\label{tab:mscoco-ITR}
\end{table*}

\begin{table*}[!t]
\setlength{\abovecaptionskip}{2pt}
\caption{\textbf{Cross-Task Transferability.} 
The Baseline represents the original performance of IC and VG on clean data. We utilize ALBEF to generate multi-modal adversarial examples for attacking both Visual Grounding (VG) and Image Captioning (IC).}
\begin{center}
\normalsize
\renewcommand\arraystretch{1}
% TODO 添加阴影之后，会超出单个格子的范围
\setlength{\tabcolsep}{16pt}
    \resizebox{\linewidth}{!}{
		% \scalebox{1.1}[1.0]{
		\begin{tabular}{ @{\extracolsep{\fill}} l|ccc|ccccc} 
        \toprule[0.3mm]
			& \multicolumn{3}{c}{\textbf{ITR $\rightarrow$ VG}} & \multicolumn{5}{c}{\textbf{ITR $\rightarrow$ IC}} \\
			\cmidrule{2-9}
			\multirow{-2}{*}{\textbf{Attack}} & {Val $\downarrow$} & {TestA $\downarrow$} &    {TestB $\downarrow$} & {B@4 $\downarrow$} & {METEOR $\downarrow$} & {ROUGE-L $\downarrow$} & {CIDEr $\downarrow$} & {SPICE $\downarrow$} \\
			\midrule
           Baseline & 58.46 & 65.89 & 46.25& 39.7 & 31.0 & 60.0 & 133.3 & 23.8 \\
Co-Attack & 54.26 & 61.80 & 43.81 & 37.4 & 29.8 & 58.4 & 125.5 & 22.8 \\
SGA & 53.55 & 61.19 & 43.71 & 34.8 & 28.4 & 56.3 & 116.0 & 21.4 \\
DRA & 53.88 & 61.18 & 43.38 & 34.8 & 28.4 & 56.4 & 115.9 & 21.4 \\
\textbf{HQA-VLAttack} & \textbf{46.48} & \textbf{54.31} & \textbf{36.90} & \textbf{31.8} & \textbf{26.8} & \textbf{54.1} & \textbf{104.6} & \textbf{19.8} \\
			\bottomrule[0.3mm]
	\end{tabular}}
\end{center}
\vspace{-20pt}
\label{tab:cross_task}
\end{table*}

\textbf{Evaluation Metrics.} We use the Attack Success Rate (ASR) as the primary evaluation metric to assess the transferability of adversarial attacks in both white-box and black-box settings. The ASR reflects the overall success rate of the attacks, with a higher ASR indicating a higher quality attack method. Additionally, we employ IR R@k, which measures the proportion of cases where none of the top-k image retrieval results contain the correct image. Similarly, we use TR R@k to represent the proportion of instances where none of the top-k caption retrieval results include the correct matching caption.

\textbf{Implementation Details.} In our experiments, we adopt adversarial attack settings of SGA in order to ensure the fairness of the comparison. For image attacks, we employ PGD with perturbation bound $\epsilon_v = 2/255$, step size $\alpha = 0.5/255$, and iteration steps $N = 10$. We leverage a combination of BERT-Attack and counter-filter word vectors to craft adversarial texts. The perturbation boundary is set to $\epsilon_t = 1$. For BERT-Attack, the length of the word list is $W = 10$. For the word vectors, the similarity threshold is set to $\tau=0.4$. In contrastive learning, the positive pair penalty factor $\lambda$ is set to $-10$, batch size is set to 16. Image scale sets $S=\{0.50, 0.75, 1.00, 1.25, 1.50\}$. Similarly, the caption set is enlarged by augmenting the most matching caption pairs for each image in the dataset, with the size of approximately five.

\subsection{Experimental Results}
\subsubsection{Image-Text Retrieval Comparison}
Our experiments focus on the Image-Text Retrieval (ITR) task, where we generate adversarial examples across various surrogate models and evaluate the effectiveness of our method using the Attack Success Rate (ASR) for both white-box and transfer attacks. As shown in Table~\ref{tab:flickr-ITR} and Table~\ref{tab:mscoco-ITR}, our approach outperforms state-of-the-art methods in ASR on both Flickr30K and MSCOCO datasets. Specifically, our method achieves nearly 100\% ASR in white-box attacks, with both Text Retrieval (TR) and Image Retrieval (IR) ASR reaching 100\% when attacking $\mathrm{CLIP_{ViT}}$ on both Flickr30K and MSCOCO. 

In black-box attacks using models with the same architecture, attacking TCL with ALBEF as the surrogate model results in ASR increases of 23.28\% (TR) and 18.77\% (IR) on Flickr30K, as well as 25.00\% (TR) and 20.35\% (IR) on MSCOCO. Similarly, when $\mathrm{CLIP_{ViT}}$ is used as the surrogate model to attack $\mathrm{CLIP_{CNN}}$, ASR gains of 28.60\% (TR) and 26.45\% (IR) on Flickr30K, with corresponding gains of 24.52\% (TR) and 19.27\% (IR) on MSCOCO. 

For attacks involving different model architectures, using ALBEF as the surrogate model to attack $\mathrm{CLIP_{ViT}}$ results in ASR improvements of 13.01\% (TR) and 13.66\% (IR) on Flickr30K, as well as 13.66\% (TR) and 12.77\% (IR) on MSCOCO. Conversely, using $\mathrm{CLIP_{ViT}}$ as the surrogate model to attack ALBEF yields ASR enhancements of 12.62\% (TR) and 11.98\% (IR) on Flickr30K, with corresponding enhancements of 20.09\% (TR) and 18.03\% (IR) on MSCOCO. 

These results demonstrate that HQA-VLAttack is a high quality attack method, significantly outperforming other approaches in terms of ASR.

\subsubsection{Cross-Task ASR Comparison}
To verify that HQA-VLAttack is not only effective in Image-Text Retrieval but also in other tasks, we conduct experiments on Image Captioning and Visual Grounding. These tasks demand strong cross-modal interaction and alignments, which are core components of multimodal learning.

\textbf{Image Captioning.} Image captioning is a generative task where the model \cite{ic_model1,ic_model2,ic_model3} first encodes the input image and then generates the corresponding textual description using a decoder based on the encoded image features. In our experiment, we select ALBEF as the surrogate model to generate adversarial examples for the Image-Text Retrieval (ITR) task and use BLIP as the victim model for image captioning. The experiment is conducted on the MSCOCO dataset, and the generated captions are evaluated using the following metrics: BLEU-4 (B@4) \cite{Bleu}, METEOR \cite{METEOR}, ROUGE \cite{Rouge}, CIDEr \cite{CIDEr}, and SPICE \cite{SPICE}. The results are shown in the Table~\ref{tab:cross_task}. It can be seen that compared with the second-best results, HQA-VLAttack improves the BLEU score by up to 3.0\% and the CIDEr score by up to 11.3\%.

\textbf{Visual Grounding.} The task of visual grounding aims to locate the region in the image that corresponds to a specific textual description. In our experiment, we use ALBEF as the surrogate model to generate adversarial examples for the Image-Text Retrieval (ITR) task and select the ALBEF model fine-tuned for visual grounding as the victim model. The experiment is conducted using the RefCOCO+ dataset, and we employ Val, TestA, and TestB as evaluation metrics. As shown in Table~\ref{tab:cross_task}, HQA-VLAttack outperforms other methods significantly.

\begin{figure*}[!t]
\begin{center}
   \includegraphics[width=0.9\linewidth]{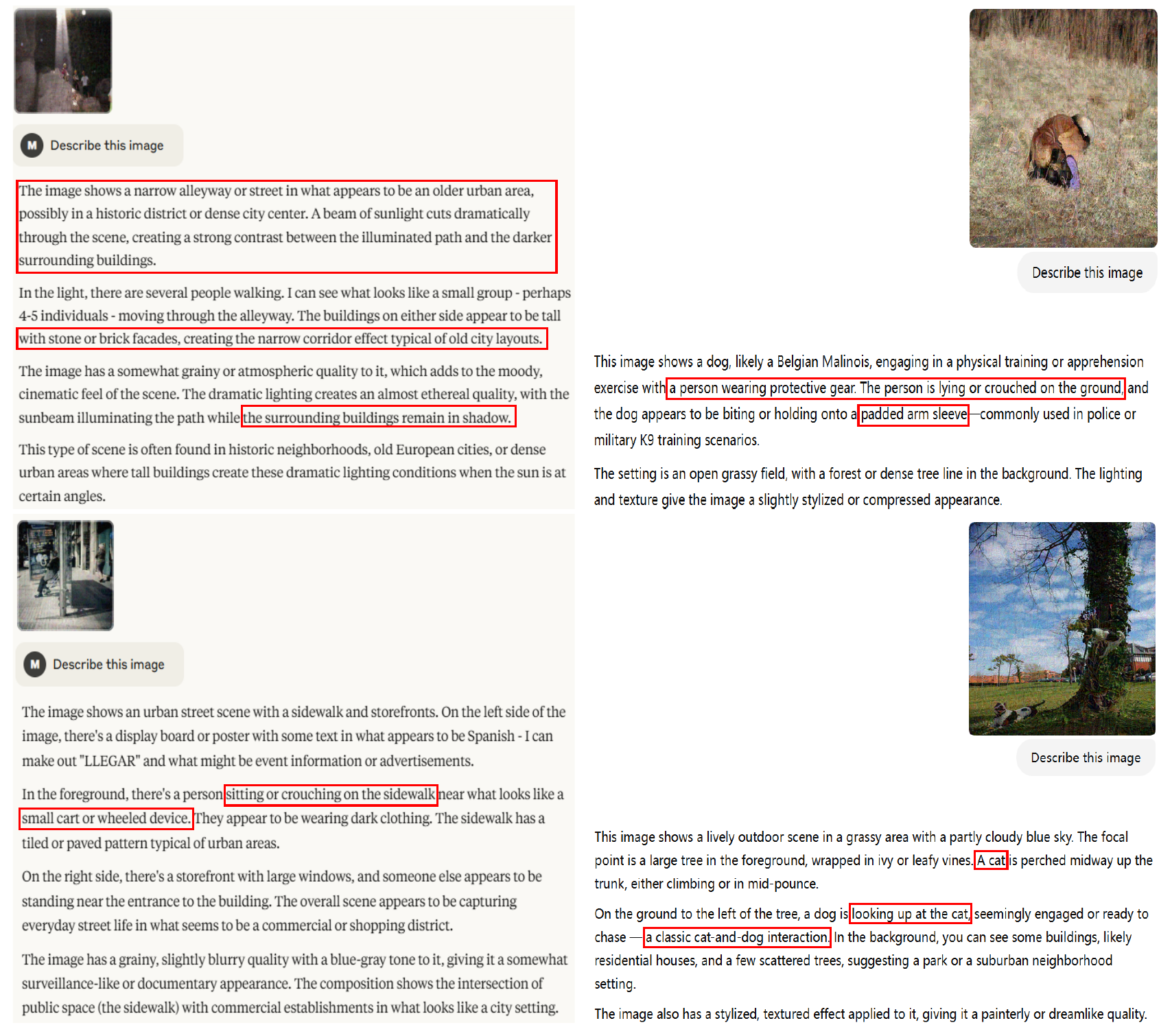}
\end{center}
   \caption{Adversarial Transferability between GPT-4o and Claude-3.7 Sonnet. The images on the left show the responses generated by Claude-3.7 Sonnet when provided with adversarial images and the prompt ``Describe this image'', while the images on the right display the outputs produced by GPT-4o under the same prompt.}
\vspace{-10pt}
\label{fig:main-mllm}
\end{figure*}

We further conduct experiments to assess the adversarial transferability of our method on Multimodal Large Language Models (MLLMs)\cite{gpt-4}. Figure~\ref{fig:main-mllm} shows adversarial examples generated by our method that successfully mislead state-of-the-art closed-source MLLMs to produce incorrect responses. Detailed experimental settings as well as additional experiments are provided in Appendix G.

\subsubsection{Ablation Study and Parameter Investigation}
We conduct experiments on ITR to evaluate the effects of different modules and batch sizes $B$ in the proposed HQA-VLAttack. Specifically, we use the Flickr30K dataset, with ALBEF as the surrogate model, and employ the Attack Success Rate (ASR, \%) as the metric. We also investigate the impact of the penalty factor $\lambda$ in Appendix F.

\begin{wrapfigure}{r}{0.55\linewidth} % 设置图形宽度为页面宽度的一半

  \centering
  \begin{subfigure}{0.48\linewidth}
    \includegraphics[width=\linewidth]{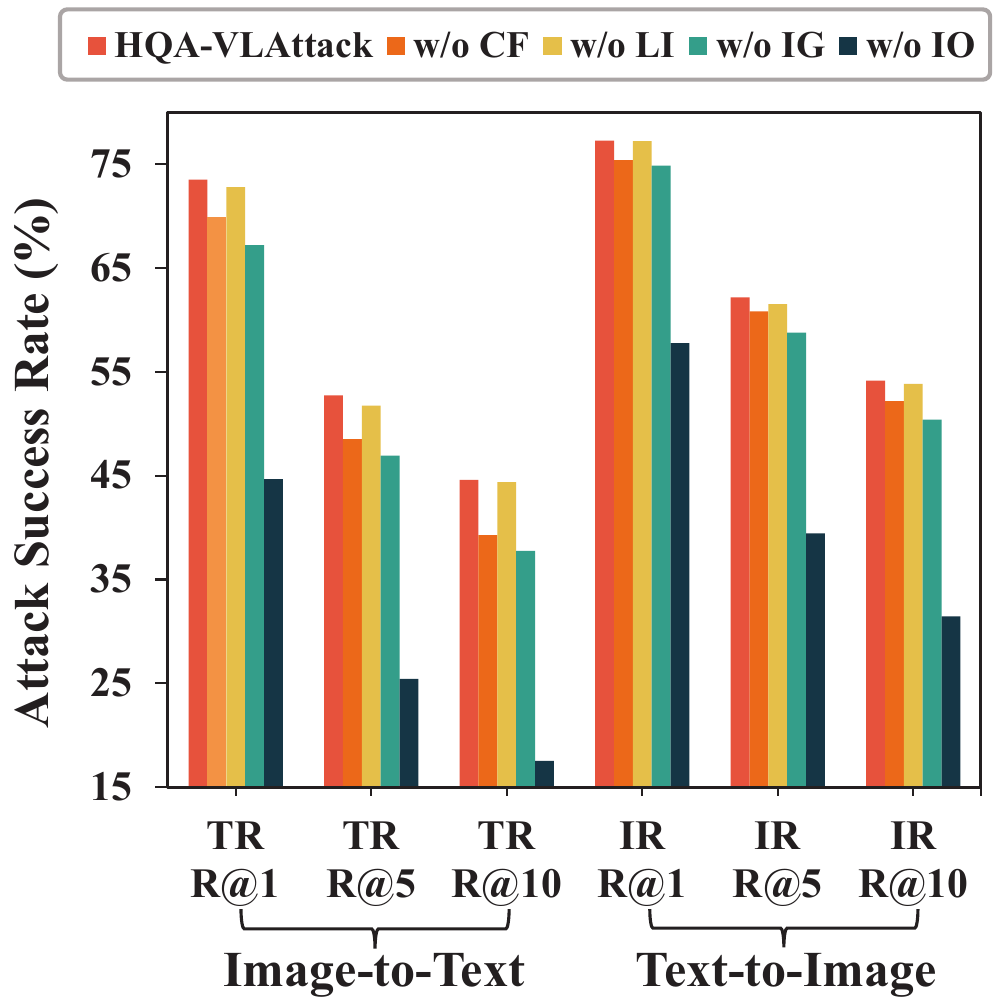} % 替换为实际图片路径
    \caption{\scriptsize Component Ablation Analysis}
    \label{fig_e:ablation-a}
  \end{subfigure}
  \hfill
  \begin{subfigure}{0.48\linewidth}
    \includegraphics[width=\linewidth]{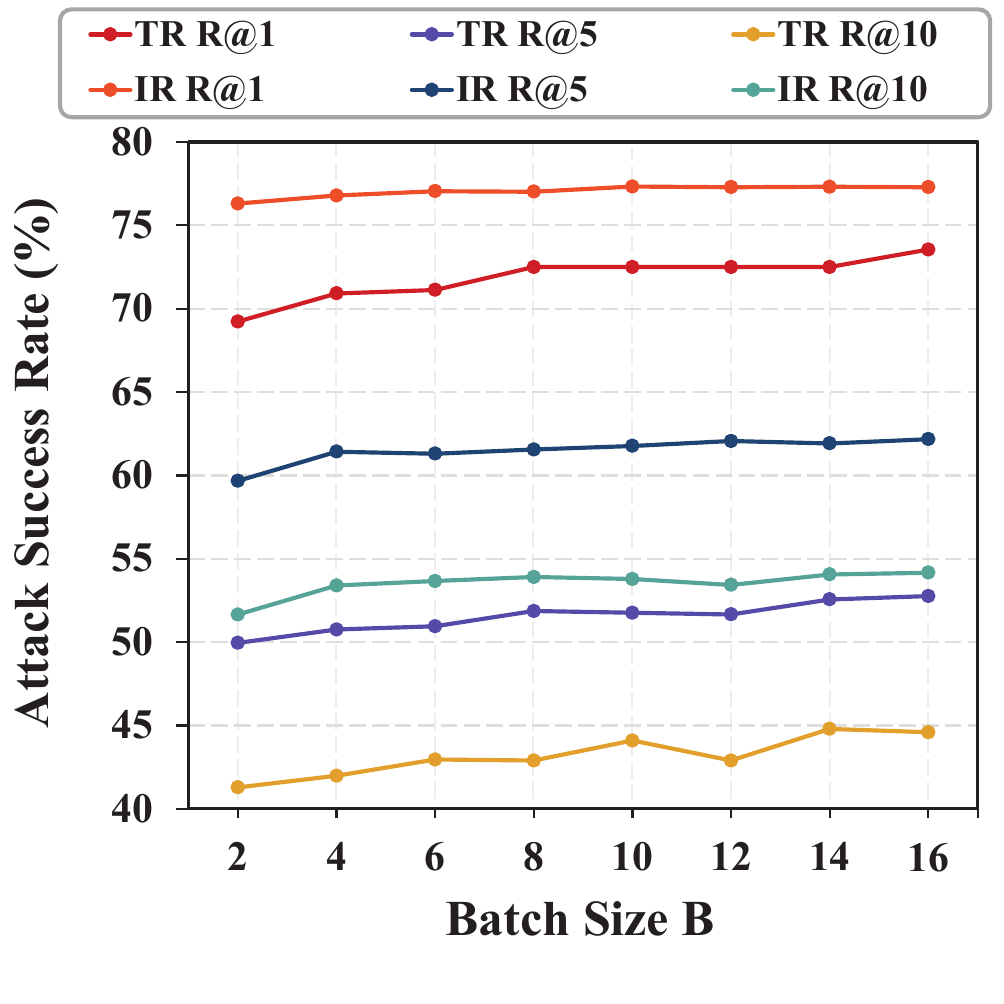} % 替换为实际图片路径
    \caption{\scriptsize Impact of Batch Size $B$ on ASR}
    \label{fig_e:ablation-b}
  \end{subfigure}
  \caption{Ablation Study on Component Effectiveness and Batch Size Impact on Attack Success Rate. }
  \vspace{-10pt}
  \label{fig_e:ablation}
\end{wrapfigure}

\textbf{Module Ablation Experiment.} When investigating the effectiveness of different components, we select TCL as the victim model. The results are shown in the Figure~\ref{fig_e:ablation-a}. 
"w/o CF" refers to the omission of the counter-fitting word vector for generating substitute words during the determining the substitute word set phase.
"w/o LI" refers to the case where no layer importance is used during the layer-importance based initial image adversarial example generation phase, and each $w_{i,l}$ is set to 1. 
"w/o IG" refers to the omission of the Layer-Importance Based Initial Image Adversarial Example Generation phase in the image attack process. 
"w/o IO" refers to the omission of the Contrastive Learning Based Image Adversarial Example Optimization. 
It is evident that all components contribute to the HQA-VLAttack's performance, validating the effectiveness of each proposed component.

\textbf{Batch Size $B$.} To investigate the impact of different batch sizes $B$ on attack success rate, we conduct an ablation analysis, and the results are shown in Figure~\ref{fig_e:ablation-b}. As $B$ increases, the contrastive learning process incorporates more unmatched texts, which helps guide the generation of adversarial images, leading to a noticeable increase in the attack success rate. These results highlight the effectiveness of a larger batch size in improving adversarial transferability. However, to balance attack performance with computational efficiency, we ultimately selected $B=16$ as the optimal batch size.

\section{Conclusion}
In this paper, we propose a novel method, HQA-VLAttack, to achieve high-quality adversarial attacks against Vision-Language Pre-training (VLP) models. For text attack, HQA-VLAttack ensures semantic consistency between the substitute word and the original word by utilizing counter-fitting word vectors to identify an appropriate set of substitute words. For image attack, HQA-VLAttack generates the initial image using a layer-importance-based approach, minimizing the similarity between the initial and original images. Moreover, by contrastive learning-based optimization, HQA-VLAttack reduces the similarity between positive image-text pairs while enhancing the similarity between negative image-text pairs. This forces the victim model to prioritize the retrieval of negative image-text pairs. Extensive experimental results demonstrate that HQA-VLAttack significantly outperforms strong baselines without requiring queries to the victim model, underscoring its effectiveness as a high-quality attack. In future work, we aim to explore further optimization strategies to refine the model and enhance its adversarial attack performance. 

\newpage

\begin{ack}
This work was supported by National Natural Science Foundation of China (No. 62206038, 62106035), Liaoning Binhai Laboratory Project (No. LBLF-2023-01), Chunhui Project Foundation of the Education Department of China (No. HZKY20220419), and Xiaomi Young Talents Program.
\end{ack}

{
    \small 
    \bibliographystyle{plain}
    \bibliography{main}

@String(CVPR= {IEEE Conf. Comput. Vis. Pattern Recog.})

@String(ICCV= {Int. Conf. Comput. Vis.})

@String(ECCV= {Eur. Conf. Comput. Vis.})

@String(TMM  = {IEEE Trans. Multimedia})

@String(ICLR = {Int. Conf. Learn. Represent.})

@String(IJCAI = {IJCAI})

@String(AAAI = {AAAI})

@String(CVPR  = {CVPR})

@String(ICCV  = {ICCV})

@String(ECCV  = {ECCV})

@String(TMM   =	{IEEE TMM})

@String(ICLR  = {ICLR})

@inproceedings{ALBEF,
  author       = {Junnan Li and
                  Ramprasaath R. Selvaraju and
                  Akhilesh Gotmare and
                  Shafiq R. Joty and
                  Caiming Xiong and
                  Steven Chu{-}Hong Hoi},
  title        = {Align before Fuse: Vision and Language Representation Learning with
                  Momentum Distillation},
  booktitle    = {Conference on Neural Information Processing Systems (NeurIPS)},
  pages        = {9694--9705},
  year         = {2021}
}

@inproceedings{CLIP,
  author       = {Alec Radford and
                  Jong Wook Kim and
                  Chris Hallacy and
                  Aditya Ramesh and
                  Gabriel Goh and
                  Sandhini Agarwal and
                  Girish Sastry and
                  Amanda Askell and
                  Pamela Mishkin and
                  Jack Clark and
                  Gretchen Krueger and
                  Ilya Sutskever},
  title        = {Learning Transferable Visual Models From Natural Language Supervision},
  booktitle    = {International Conference on Machine Learning (ICML)},
  pages        = {8748--8763},
  year         = {2021}
}

@inproceedings{TCL,
  author       = {Jinyu Yang and
                  Jiali Duan and
                  Son Tran and
                  Yi Xu and
                  Sampath Chanda and
                  Liqun Chen and
                  Belinda Zeng and
                  Trishul Chilimbi and
                  Junzhou Huang},
  title        = {Vision-Language Pre-Training with Triple Contrastive Learning},
  booktitle    = {IEEE/CVF Conference on Computer Vision and Pattern Recognition (CVPR)},
  pages        = {15650--15659},
  year         = {2022}
}

@inproceedings{HQA-Attack,
  author       = {Han Liu and
                  Zhi Xu and
                  Xiaotong Zhang and
                  Feng Zhang and
                  Fenglong Ma and
                  Hongyang Chen and
                  Hong Yu and
                  Xianchao Zhang},
  title        = {HQA-Attack: Toward High Quality Black-Box Hard-Label Adversarial Attack
                  on Text},
  booktitle    = {Conference on Neural Information Processing Systems (NeurIPS)},
  year         = {2023}
}

@inproceedings{SSPAttack,
  author       = {Han Liu and
                  Zhi Xu and
                  Xiaotong Zhang and
                  Xiaoming Xu and
                  Feng Zhang and
                  Fenglong Ma and
                  Hongyang Chen and
                  Hong Yu and
                  Xianchao Zhang},
  title        = {SSPAttack: {A} Simple and Sweet Paradigm for Black-Box Hard-Label
                  Textual Adversarial Attack},
  booktitle    = {AAAI Conference on Artificial Intelligence (AAAI)},
  pages        = {13228--13235},
  year         = {2023}
}

@article{gao2023adversarial,
  title={Adversarial Neural Collaborative Filtering with Embedding Dimension Correlations},
  author={Gao, Yi and Chen, Jianxia and Xiao, Liang and Wang, Hongyang and Pan, Liwei and Wen, Xuan and Ye, Zhiwei and Wu, Xinyun},
  journal={Data Intelligence},
  volume={5},
  number={3},
  pages={786--806},
  year={2023},
  publisher={China Science Publishing & Media Ltd.},
  doi={10.1162/dint_a_00151}
}

@inproceedings{DRA,
  author       = {Sensen Gao and
                  Xiaojun Jia and
                  Xuhong Ren and
                  Ivor W. Tsang and
                  Qing Guo},
  title        = {Boosting Transferability in Vision-Language Attacks via Diversification
                  Along the Intersection Region of Adversarial Trajectory},
  booktitle    = {European Conference on Computer Vision (ECCV)},
  pages        = {442--460},
  year         = {2024}
}

@inproceedings{SGA,
  author       = {Dong Lu and
                  Zhiqiang Wang and
                  Teng Wang and
                  Weili Guan and
                  Hongchang Gao and
                  Feng Zheng},
  title        = {Set-level Guidance Attack: Boosting Adversarial Transferability of
                  Vision-Language Pre-training Models},
  booktitle    = {IEEE International Conference on Computer Vision (ICCV)},
  pages        = {102--111},
  year         = {2023}
}

@inproceedings{PGD,
  author       = {Aleksander Madry and
                  Aleksandar Makelov and
                  Ludwig Schmidt and
                  Dimitris Tsipras and
                  Adrian Vladu},
  title        = {Towards Deep Learning Models Resistant to Adversarial Attacks},
  booktitle    = {International Conference on Learning Representations (ICLR)},
  year         = {2018}
}

@inproceedings{BERT-ATTACK,
  author       = {Linyang Li and
                  Ruotian Ma and
                  Qipeng Guo and
                  Xiangyang Xue and
                  Xipeng Qiu},
  title        = {{BERT-ATTACK:} Adversarial Attack Against {BERT} Using {BERT}},
  booktitle    = {Conference on Empirical Methods in Natural Language Processing (EMNLP)},
  pages        = {6193--6202},
  year         = {2020}
}

@inproceedings{Co-Attack,
  author       = {Jiaming Zhang and
                  Qi Yi and
                  Jitao Sang},
  title        = {Towards Adversarial Attack on Vision-Language Pre-training Models},
  booktitle    = {ACM International Conference on Multimedia (MM)},
  pages        = {5005--5013},
  year         = {2022}
}

@article{IC,
  author       = {Matteo Stefanini and
                  Marcella Cornia and
                  Lorenzo Baraldi and
                  Silvia Cascianelli and
                  Giuseppe Fiameni and
                  Rita Cucchiara},
  title        = {From Show to Tell: {A} Survey on Deep Learning-Based Image Captioning},
  journal      = {{IEEE} Trans. Pattern Anal. Mach. Intell.},
  volume       = {45},
  number       = {1},
  pages        = {539--559},
  year         = {2023}
}

@inproceedings{Refcoco+,
  author       = {Licheng Yu and
                  Patrick Poirson and
                  Shan Yang and
                  Alexander C. Berg and
                  Tamara L. Berg},
  title        = {Modeling Context in Referring Expressions},
  booktitle    = {European Conference on Computer Vision (ECCV)},
  pages        = {69--85},
  year         = {2016}
}

@inproceedings{SPICE,
  author       = {Peter Anderson and
                  Basura Fernando and
                  Mark Johnson and
                  Stephen Gould},
  title        = {{SPICE:} Semantic Propositional Image Caption Evaluation},
  booktitle    = {European Conference on Computer Vision (ECCV)},
  pages        = {382--398},
  year         = {2016}
}

@inproceedings{METEOR,
  author       = {Satanjeev Banerjee and
                  Alon Lavie},
  title        = {{METEOR:} An Automatic Metric for {MT} Evaluation with Improved Correlation
                  with Human Judgments},
  booktitle    = {Proceedings of the Workshop on Intrinsic and Extrinsic Evaluation
                  Measures for Machine Translation and/or Summarization},
  pages        = {65--72},
  year         = {2005}
}

@inproceedings{Rouge,
  title={Rouge: A package for automatic evaluation of summaries},
  author={Chin-Yew Lin},
  booktitle={In Annual Meeting of the Association for Computational Linguistics},
  year={2004}
}

@inproceedings{Bleu,
  author       = {Kishore Papineni and
                  Salim Roukos and
                  Todd Ward and
                  Wei{-}Jing Zhu},
  title        = {Bleu: a Method for Automatic Evaluation of Machine Translation},
  booktitle    = {Annual Meeting of the Association for Computational Linguistics (ACL)},
  pages        = {311--318},
  year         = {2002}
}

@inproceedings{CIDEr,
  author       = {Ramakrishna Vedantam and
                  C. Lawrence Zitnick and
                  Devi Parikh},
  title        = {CIDEr: Consensus-based image description evaluation},
  booktitle    = {IEEE/CVF Conference on Computer Vision and Pattern Recognition (CVPR)},
  pages        = {4566--4575},
  year         = {2015}
}

@inproceedings{itr1,
  author       = {Min Cao and
                  Shiping Li and
                  Juntao Li and
                  Liqiang Nie and
                  Min Zhang},
  title        = {Image-text Retrieval: {A} Survey on Recent Research and Development},
  booktitle    = {International Joint Conference on Artificial Intelligence (IJCAI)},
  pages        = {5410--5417},
  year         = {2022}
}

@inproceedings{white1_vl,
  author       = {Xiaojun Xu and
                  Xinyun Chen and
                  Chang Liu and
                  Anna Rohrbach and
                  Trevor Darrell and
                  Dawn Song},
  title        = {Fooling Vision and Language Models Despite Localization and Attention
                  Mechanism},
  booktitle    = {IEEE/CVF Conference on Computer Vision and Pattern Recognition (CVPR)},
  pages        = {4951--4961},
  year         = {2018},
}

@inproceedings{VLAttack,
  author       = {Ziyi Yin and
                  Muchao Ye and
                  Tianrong Zhang and
                  Tianyu Du and
                  Jinguo Zhu and
                  Han Liu and
                  Jinghui Chen and
                  Ting Wang and
                  Fenglong Ma},
  title        = {{VLATTACK:} Multimodal Adversarial Attacks on Vision-Language Tasks
                  via Pre-trained Models},
  booktitle    = {Conference on Neural Information Processing Systems (NeurIPS)},
  year         = {2023}
}

@inproceedings{MSCOCO,
  author       = {Tsung{-}Yi Lin and
                  Michael Maire and
                  Serge J. Belongie and
                  James Hays and
                  Pietro Perona and
                  Deva Ramanan and
                  Piotr Doll{\'{a}}r and
                  C. Lawrence Zitnick},
  title        = {Microsoft {COCO:} Common Objects in Context},
  booktitle    = {European Conference on Computer Vision (ECCV)},
  pages        = {740--755},
  year         = {2014}
}

@inproceedings{UAP,
  author       = {Peng{-}Fei Zhang and
                  Zi Huang and
                  Guangdong Bai},
  title        = {Universal Adversarial Perturbations for Vision-Language Pre-trained
                  Models},
  booktitle    = {Annual International ACM SIGIR Conference on Research and Development in Information Retrieval (SIGIR)},
  pages        = {862--871},
  year         = {2024}
}

@inproceedings{ID,
  author       = {Cihang Xie and
                  Zhishuai Zhang and
                  Yuyin Zhou and
                  Song Bai and
                  Jianyu Wang and
                  Zhou Ren and
                  Alan L. Yuille},
  title        = {Improving Transferability of Adversarial Examples With Input Diversity},
  booktitle    = {IEEE/CVF Conference on Computer Vision and Pattern Recognition (CVPR)},
  pages        = {2730--2739},
  year         = {2019}
}

@inproceedings{ViT,
  author       = {Alexey Dosovitskiy and
                  Lucas Beyer and
                  Alexander Kolesnikov and
                  Dirk Weissenborn and
                  Xiaohua Zhai and
                  Thomas Unterthiner and
                  Mostafa Dehghani and
                  Matthias Minderer and
                  Georg Heigold and
                  Sylvain Gelly and
                  Jakob Uszkoreit and
                  Neil Houlsby},
  title        = {An Image is Worth 16x16 Words: Transformers for Image Recognition
                  at Scale},
  booktitle    = {International Conference on Learning Representations (ICLR)},
  year         = {2021}
}

@inproceedings{Flickr30k,
  author       = {Bryan A. Plummer and
                  Liwei Wang and
                  Chris M. Cervantes and
                  Juan C. Caicedo and
                  Julia Hockenmaier and
                  Svetlana Lazebnik},
  title        = {Flickr30k Entities: Collecting Region-to-Phrase Correspondences for
                  Richer Image-to-Sentence Models},
  booktitle    = {IEEE International Conference on Computer Vision (ICCV)},
  pages        = {2641--2649},
  year         = {2015}
}

@article{itr2,
  author       = {Jiamin Zhuang and
                  Jing Yu and
                  Yang Ding and
                  Xiangyan Qu and
                  Yue Hu},
  title        = {Towards Fast and Accurate Image-Text Retrieval With Self-Supervised
                  Fine-Grained Alignment},
  journal      = {IEEE Transactions on Multimedia (TMM)},
  volume       = {26},
  pages        = {1361--1372},
  year         = {2024}
}

@inproceedings{resnet,
  author       = {Kaiming He and
                  Xiangyu Zhang and
                  Shaoqing Ren and
                  Jian Sun},
  title        = {Deep Residual Learning for Image Recognition},
  booktitle    = {IEEE/CVF Conference on Computer Vision and Pattern Recognition (CVPR)},
  pages        = {770--778},
  year         = {2016}
}

@inproceedings{DBLP:conf/eccv/MaJWYQ24,
  author       = {Chuofan Ma and
                  Yi Jiang and
                  Jiannan Wu and
                  Zehuan Yuan and
                  Xiaojuan Qi},
  title        = {Groma: Localized Visual Tokenization for Grounding Multimodal Large
                  Language Models},
  booktitle    = {European Conference on Computer Vision (ECCV)},
  pages        = {417--435},
  year         = {2024}
}

@article{multimodalattack1,
  author       = {Hao Cheng and
                  Erjia Xiao and
                  Jiahang Cao and
                  Le Yang and
                  Kaidi Xu and
                  Jindong Gu and
                  Renjing Xu},
  title        = {Typography Leads Semantic Diversifying: Amplifying Adversarial Transferability
                  across Multimodal Large Language Models},
  journal      = {arXiv preprint},
  archiveprefix  = {arXiv},
  volume       = {abs/2405.20090},
  year         = {2024}
}

@article{black2_q,
  author       = {Zhaoyu Chen and
                  Bo Li and
                  Shuang Wu and
                  Shouhong Ding and
                  Wenqiang Zhang},
  title        = {Query-Efficient Decision-Based Black-Box Patch Attack},
  journal      = {IEEE Transactions on Information Forensics and Security},
  volume       = {18},
  pages        = {5522--5536},
  year         = {2023}
}

@article{itr3,
  author       = {Zheng Cui and
                  Yongli Hu and
                  Yanfeng Sun and
                  Junbin Gao and
                  Baocai Yin},
  title        = {Cross-modal alignment with graph reasoning for image-text retrieval},
  journal      = {Multimedia Tools and Applications},
  volume       = {81},
  number       = {17},
  pages        = {23615--23632},
  year         = {2022}
}

@inproceedings{bert_mlm,
  author       = {Jacob Devlin and
                  Ming{-}Wei Chang and
                  Kenton Lee and
                  Kristina Toutanova},
  title        = {{BERT:} Pre-training of Deep Bidirectional Transformers for Language
                  Understanding},
  booktitle    = {North American Chapter of the Association for Computational Linguistics (NAACL)},
  year         = {2019}
}

@inproceedings{VQAttack,
  author       = {Ziyi Yin and
                  Muchao Ye and
                  Tianrong Zhang and
                  Jiaqi Wang and
                  Han Liu and
                  Jinghui Chen and
                  Ting Wang and
                  Fenglong Ma},
  title        = {VQAttack: Transferable Adversarial Attacks on Visual Question Answering
                  via Pre-trained Models},
  booktitle    = {AAAI Conference on Artificial Intelligence (AAAI)},
  year         = {2024}
}

@inproceedings{counter-fitting,
  author       = {Nikola Mrksic and
                  Diarmuid {\'{O}} S{\'{e}}aghdha and
                  Blaise Thomson and
                  Milica Gasic and
                  Lina Maria Rojas{-}Barahona and
                  Pei{-}Hao Su and
                  David Vandyke and
                  Tsung{-}Hsien Wen and
                  Steve J. Young},
  title        = {Counter-fitting Word Vectors to Linguistic Constraints},
  booktitle    = {North American Chapter of the Association for Computational Linguistics (NAACL)},
  year         = {2016}
}

@article{ic_model1,
  author       = {Jie Wu and
                  Tianshui Chen and
                  Hefeng Wu and
                  Zhi Yang and
                  Guangchun Luo and
                  Liang Lin},
  title        = {Fine-Grained Image Captioning With Global-Local Discriminative Objective},
  journal      = {IEEE Transactions on Multimedia (TMM)},
  volume       = {23},
  pages        = {2413--2427},
  year         = {2021}
}

@article{ic_model2,
  author       = {Depeng Wang and
                  Zhenzhen Hu and
                  Yuanen Zhou and
                  Richang Hong and
                  Meng Wang},
  title        = {A Text-Guided Generation and Refinement Model for Image Captioning},
  journal      = {IEEE Transactions on Multimedia (TMM)},
  volume       = {25},
  pages        = {2966--2977},
  year         = {2023}
}

@article{ic_model3,
  author       = {Majjed Al{-}Qatf and
                  Xingfu Wang and
                  Ammar Hawbani and
                  Amr Abdussalam and
                  Saeed Hamood Alsamhi},
  title        = {Image Captioning With Novel Topics Guidance and Retrieval-Based Topics
                  Re-Weighting},
  journal      = {IEEE Transactions on Multimedia (TMM)},
  volume       = {25},
  pages        = {5984--5999},
  year         = {2023}
}

@article{gpt-4,
  author       = {OpenAI},
  title        = {{GPT-4} Technical Report},
  journal      = {arXiv preprint},
  archiveprefix  = {arXiv},
  volume       = {abs/2303.08774},
  year         = {2023}
}
}

\newpage

\section*{NeurIPS Paper Checklist}

\begin{enumerate}

\item {\bf Claims}
    \item[] Question: Do the main claims made in the abstract and introduction accurately reflect the paper's contributions and scope?
    \item[] Answer: \answerYes{}% Replace by \answerYes{}, \answerNo{}, or \answerNA{}.
    \item[] Justification: The abstract and introduction provide a comprehensive outline of the motivations and contributions of the paper, which are subsequently examined and validated through thorough experimentation.
    
\item {\bf Limitations}
    \item[] Question: Does the paper discuss the limitations of the work performed by the authors?
    \item[] Answer:  \answerYes{} % Replace by \answerYes{}, \answerNo{}, or \answerNA{}.
    \item[] Justification: We discuss the limitations of our method in Appendix I
    
\item {\bf Theory assumptions and proofs}
    \item[] Question: For each theoretical result, does the paper provide the full set of assumptions and a complete (and correct) proof?
    \item[] Answer: \answerNA{} % Replace by \answerYes{}, \answerNo{}, or \answerNA{}.
    \item[] Justification: This paper does not involve extensive theoretical assumptions.

    \item {\bf Experimental result reproducibility}
    \item[] Question: Does the paper fully disclose all the information needed to reproduce the main experimental results of the paper to the extent that it affects the main claims and/or conclusions of the paper (regardless of whether the code and data are provided or not)?
    \item[] Answer: \answerYes{} % Replace by \answerYes{}, \answerNo{}, or \answerNA{}.
    \item[] Justification: We provide all the hyperparameter settings in the experimental settings and Appendix F. Additionally, we submit our code in the supplementary materials.

\item {\bf Open access to data and code}
    \item[] Question: Does the paper provide open access to the data and code, with sufficient instructions to faithfully reproduce the main experimental results, as described in supplemental material?
    \item[] Answer: \answerYes{} % Replace by \answerYes{}, \answerNo{}, or \answerNA{}.
    \item[] Justification: All data utilized in this study are sourced from open-source platforms, ensuring their openness, transparency, and accessibility. We submit our code and execution scripts as supplementary materials.

\item {\bf Experimental setting/details}
    \item[] Question: Does the paper specify all the training and test details (e.g., data splits, hyperparameters, how they were chosen, type of optimizer, etc.) necessary to understand the results?
    \item[] Answer: \answerYes{} % Replace by \answerYes{}, \answerNo{}, or \answerNA{}.
    \item[] Justification: We provide all the hyperparameter settings in the experimental settings and Appendix F. 

\item {\bf Experiment statistical significance}
    \item[] Question: Does the paper report error bars suitably and correctly defined or other appropriate information about the statistical significance of the experiments?
    \item[] Answer: \answerNo{} % Replace by \answerYes{}, \answerNo{}, or \answerNA{}.
    \item[] Justification: We follow prior works and use the same random seed for reproducibility. As the results are stable and previous studies rarely report statistical significance analysis, we do not include error bars or confidence intervals.

\item {\bf Experiments compute resources}
    \item[] Question: For each experiment, does the paper provide sufficient information on the computer resources (type of compute workers, memory, time of execution) needed to reproduce the experiments?
    \item[] Answer: \answerYes{} % Replace by \answerYes{}, \answerNo{}, or \answerNA{}.
    \item[] Justification: We list the detailed computer resources In Appendix F
    
\item {\bf Code of ethics}
    \item[] Question: Does the research conducted in the paper conform, in every respect, with the NeurIPS Code of Ethics \url{https://neurips.cc/public/EthicsGuidelines}?
    \item[] Answer: \answerYes{} % Replace by \answerYes{}, \answerNo{}, or \answerNA{}.
    \item[] Justification: We rigorously adhered to the NeurIPS Code of Ethics in our research, ensuring integrity, respect for all stakeholders, compliance with ethical guidelines, and absolute anonymity.

\item {\bf Broader impacts}
    \item[] Question: Does the paper discuss both potential positive societal impacts and negative societal impacts of the work performed?
    \item[] Answer: \answerYes{} % Replace by \answerYes{}, \answerNo{}, or \answerNA{}.
    \item[] Justification: The broad impacts are detailed in Appendix J.
    
\item {\bf Safeguards}
    \item[] Question: Does the paper describe safeguards that have been put in place for responsible release of data or models that have a high risk for misuse (e.g., pretrained language models, image generators, or scraped datasets)?
    \item[] Answer: \answerNA{} % Replace by \answerYes{}, \answerNo{}, or \answerNA{}.
    \item[] Justification: Our paper presents no release of data or models with significant misuse potential, like pretrained language models or scraped datasets.

\item {\bf Licenses for existing assets}
    \item[] Question: Are the creators or original owners of assets (e.g., code, data, models), used in the paper, properly credited and are the license and terms of use explicitly mentioned and properly respected?
    \item[] Answer: \answerYes{} % Replace by \answerYes{}, \answerNo{}, or \answerNA{}.
    \item[] Justification: In this paper, we acknowledge the creators or original owners of all assets, including code, data, and models, respecting intellectual property rights.

\item {\bf New assets}
    \item[] Question: Are new assets introduced in the paper well documented and is the documentation provided alongside the assets?
    \item[] Answer: \answerYes{} % Replace by \answerYes{}, \answerNo{}, or \answerNA{}.
    \item[] Justification: 
The paper introduces new assets via code, which are thoroughly documented in the supplemental material.

\item {\bf Crowdsourcing and research with human subjects}
    \item[] Question: For crowdsourcing experiments and research with human subjects, does the paper include the full text of instructions given to participants and screenshots, if applicable, as well as details about compensation (if any)? 
    \item[] Answer: \answerNA{} % Replace by \answerYes{}, \answerNo{}, or \answerNA{}.
    \item[] Justification: The paper does not involve crowdsourcing or research with human subjects.
    
\item {\bf Institutional review board (IRB) approvals or equivalent for research with human subjects}
    \item[] Question: Does the paper describe potential risks incurred by study participants, whether such risks were disclosed to the subjects, and whether Institutional Review Board (IRB) approvals (or an equivalent approval/review based on the requirements of your country or institution) were obtained?
    \item[] Answer: \answerNA{} % Replace by \answerYes{}, \answerNo{}, or \answerNA{}.
    \item[] Justification: This paper does not utilize crowdsourcing or conduct research involving human participants.

\item {\bf Declaration of LLM usage}
    \item[] Question: Does the paper describe the usage of LLMs if it is an important, original, or non-standard component of the core methods in this research? Note that if the LLM is used only for writing, editing, or formatting purposes and does not impact the core methodology, scientific rigorousness, or originality of the research, declaration is not required.
    %this research? 
    \item[] Answer: \answerNA{} % Replace by \answerYes{}, \answerNo{}, or \answerNA{}.
    \item[] Justification: This paper utilizes LLMs solely for assessing experimental outcomes in Appendix G and they are not central to the study's main content.
    
\end{enumerate}

\end{document}